%% file: main.tex
\documentclass[10pt,letterpaper,twocolumn]{article}
\usepackage[latin9]{inputenc}
\usepackage{array}
\usepackage{booktabs}
\usepackage{mathtools}
\usepackage{multirow}
\usepackage{amsmath}
\usepackage{amssymb}
\usepackage{graphicx}
\usepackage[unicode=true,
 bookmarks=false,
 breaklinks=true,pdfborder={0 0 1},backref=page,colorlinks=false]
 {hyperref}
\hypersetup{
 pagebackref=true,letterpaper=true,colorlinks}

\makeatletter

\providecommand{\tabularnewline}{\\}

\usepackage{iccv}
\usepackage{times}
\usepackage{epsfig}
\usepackage{graphicx}

\iccvfinalcopy % *** Uncomment this line for the final submission

\ificcvfinal\fi

\usepackage[capitalize]{cleveref}
\crefname{section}{Sec.}{Secs.}
\Crefname{section}{Section}{Sections}
\Crefname{table}{Table}{Tables}
\crefname{table}{Tab.}{Tabs.}

\AtBeginDocument{%

\let\ref\cref
}

\@ifundefined{showcaptionsetup}{}{%
 \PassOptionsToPackage{caption=false}{subfig}}
\usepackage{subfig}
\makeatother

\begin{document}
\title{Persistent-Transient Duality: A Multi-mechanism Approach for Modeling
Human-Object Interaction}
\author{Hung Tran$^{1}$, Vuong Le$^{2}$, Svetha Venkatesh$^{1}$, Truyen
Tran$^{1}$\\
 {\normalsize{}$^{1}$Applied AI Institute, Deakin University, $^{2}$Amazon}\\
{\normalsize{}\{tduy, svetha.venkatesh, truyen.tran\}@deakin.edu.au,
levuong@amazon.com}}
\maketitle
\begin{abstract}
\input{abs.tex}
\end{abstract}

\section{Introduction}

\input{intro.tex}

\section{Related Work}

\input{related.tex}

\section{Method \label{sec:Method}}

\input{method.tex}

\section{Experiments \label{sec:Experiments}}

\input{exp.tex}

\section{Discussion}

\input{discuss.tex}

{\small{}\bibliographystyle{ieee_fullname}
\bibliography{egbib,hungtd,hungtd_trajectory}
 }{\small\par}
\end{document}

%% file: abs.tex
Humans are highly adaptable, swiftly switching between different modes
to progressively handle different tasks, situations and contexts.
In Human-object interaction (HOI) activities, these modes can be attributed
to two mechanisms: (1) the large-scale consistent plan for the whole
activity and (2) the small-scale children interactive actions that
start and end along the timeline. While neuroscience and cognitive
science have confirmed this multi-mechanism nature of human behavior,
machine modeling approaches for human motion are trailing behind.
While attempted to use gradually morphing structures (e.g., graph
attention networks) to model the dynamic HOI patterns, they miss the
expeditious and discrete mode-switching nature of the human motion.
To bridge that gap, this work proposes to model two concurrent mechanisms
that jointly control human motion: the Persistent process that runs
continually on the global scale, and the Transient sub-processes that
operate intermittently on the local context of the human while interacting
with objects. These two mechanisms form an interactive Persistent-Transient
Duality that synergistically governs the activity sequences. We model
this conceptual duality by a parent-child neural network of Persistent
and Transient channels with a dedicated neural module for dynamic
mechanism switching. The framework is trialed on HOI motion forecasting.
On two rich datasets and a wide variety of settings, the model consistently
delivers superior performances, proving its suitability for the challenge.

%% file: intro.tex
\begin{figure}
\begin{centering}
\includegraphics[width=0.9\columnwidth]{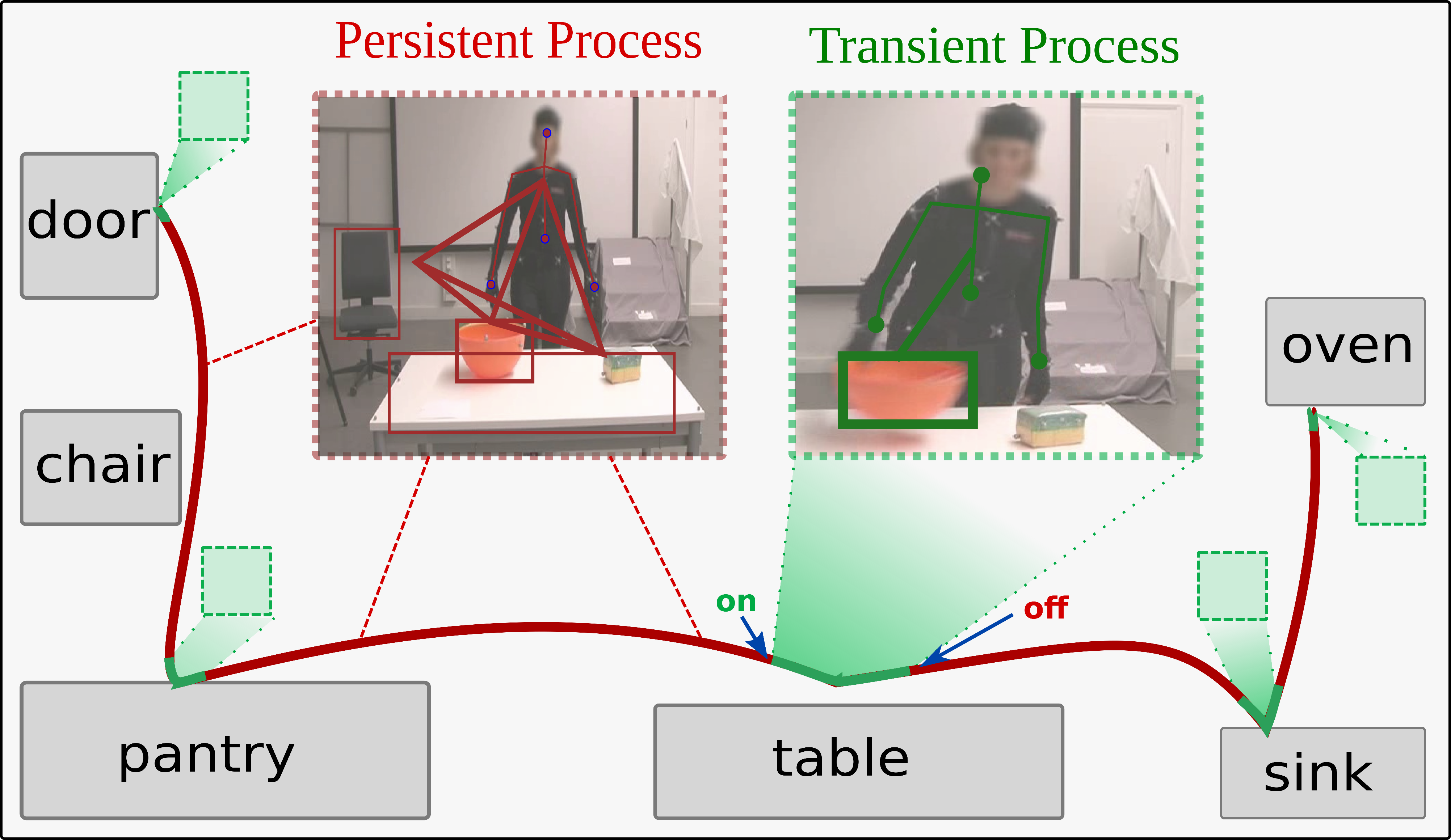}
\par\end{centering}
\caption{We model the human motions by a duality of two types of processes:
A single \emph{Persistent process} (red box) considers the whole activity
plan and trajectory (red curve) and use dense relational structure
(red graph). Its children \emph{Transient processes} (green boxes)
work on the small and local spatiotemporal scope of a human and the
interacted objects using egocentric structure (green graph). They
get turn on and off on-demand by the Persistent process.\label{fig:intro_demonstration}}
\end{figure}
In a planned activity that involves interaction with objects, the
patterns of human motion vary greatly, contextualized to the situation
and stages of the plan. The variety of patterns comes from the fact
that the subject needs to switch between multiple modes of operations:
navigating the global scene according to the plan, and occasionally
concentrate on a small set of objects to act on them. This mode-switching
nature requires a machine model to adapt quickly in structure, representation,
and inference mechanisms to follow the patterns of the behavior. This
observation is confirmed by neuroscience findings that human brain
activity contains transient networks that deals with particular situation
\cite{baker2014fast}. In cognitive science, human activities are
also proved to follow parent-child planning patterns \cite{ajzen1985intentions}.

Recent advancements in graph neural networks \cite{velivckovic2017graph}
allow motion models to dynamically adjust their relational structure
to adapt to changing situations \cite{corona2020context}. However,
with a singular inference mechanism, they can only have \emph{gradual
adaptation}: smoothly adjusting parameters of the same model. They
cannot account for quick and abrupt changes\emph{ }regarding the discrete
switching between distinctive mechanisms and as a result fail to keep
up with the movement patterns. This limitation is inherent in human-object
interaction motion (HOI-M) prediction. \ref{fig:intro_demonstration}
visualizes an example activity of cooking a dinner meal. Here the
subject navigates around the kitchen following a recipe and consider
the whole kitchen floor plan (red trajectory in \ref{fig:intro_demonstration}).
Occasionally, they deviate to perform a particular action by interacting
with an object (green sections in \ref{fig:intro_demonstration})
such as moving a bowl, where only the bowl needs their attention.
When this happens, these models will continue to consider the interacted
object as an equal member of the scene and miss its importance as
the direct recipient of the action.

To address this limitation, we explore a new modeling paradigm that
factorizes the human-object interaction into two internal types of
processes: a \emph{Persistent process} (red box in \ref{fig:intro_demonstration})
which maintains a continuous large-scale activity progress; and\emph{
Transient sub-processes} (green boxes in \ref{fig:intro_demonstration})
which has an adaptive life cycle and a personalized structure to reflects
the small-scale local interaction with objects. The Transient act
as Persistent\textquoteright s sub-processes, and they can be turned
on or off by a switch (on/off arrows in \ref{fig:intro_demonstration})
operating on signal from the main Persistent process. The\emph{ parent-child
relationship} connects the two types of processes and constitute the
\emph{Persistent-Transient Duality}. 

This modeling is related to similar transient concepts in other fields
such as control theory \cite{henzinger2000theory}, electrical \cite{greenwood1991electrical},
and chemical engineering \cite{van2001transients,duncan2019chemical}.
In computer engineering, the parent-child relationship also resembles
the operating system and the children task-specific processes.

We model this concept into a multi-channel neural network called \emph{Persistent-Transient
Duality }(PTD) networks for Human-object interaction motion (HOI-M)
prediction. The \emph{Persistent channel} contains a recurrent relational
network operating on the global scene spatially and throughout the
session temporally. The \emph{Transient channels} instead have contextualized
structures constructed on the spot whenever the human subjects shift
the priority toward interacting with a subset of objects. The life
cycles of these spontaneous channels are managed by a neural \emph{Transient
Switch}, which anticipates the initialization and termination of many
Transient processes along a single Persistent process. 

The benefit of PTD is demonstrated via motion forecasting experiments
on WBHM and Bimanual Actions datasets where it sets new SOTA performances
and generalizability tests. Being a generic framework for human behavior
modeling, PTD is readily applicable to other problems such as pedestrian
trajectory prediction of which preliminary adaptation and experiment
are shown in the supplementary. 

The key contributions of this work are:

1. The exploration of \emph{the new Persistent-transient duality concept}
to model the multi-mechanism nature of human behavior reflected in
the large-small temporal scale and global-local spatial scope of HOI
motions.

2. A \emph{parent-children neural framework} with egocentric design
that applies the PTD concept in HOI-M prediction.

3. The extensive analysis demonstrating that PTD \emph{sets new SoTA
in HOI-M prediction} across multiple datasets and settings, and generalizes
better to new scenarios.

%% file: related.tex
\textbf{Human-object interaction} in videos is traditionally formulated
for predictions of human action and object affordance labels \cite{liu2020forecasting,nagarajan2019grounded,ji2021detecting,escorcia2013spatio}.
Approaches include modeling the activity continuity with CNNs \cite{liu2020forecasting}
and RNNs \cite{nagarajan2019grounded}, or modeling the interactions
using GNNs \cite{jain2016structural,ghosh2020stacked}, which can
be improved with dynamic topology \cite{morais2021learning,qi2018learning}.
Recent works also leverage the power of transformers \cite{vaswani2017attention}
to model the spatio-temporal relationship between humans and objects
in the videos \cite{ji2021detecting}. In this paper, we consider
human-object interaction motion (HOI-M) prediction, the task of forecasting
the concrete future locations of humans and objects in an activity
which is more well-defined and easier to evaluate than the vague affordance
labels traditionally used in HOI. Also, this task is more complex
and requires more precise modeling structures, such as the duality
proposed in this work.

\textbf{Human body motion} predictions are done with MLPs \cite{bouazizi2022motionmixer,guo2023back},
RNNs \cite{fragkiadaki2015recurrent,ghosh2017learning,pavllo2018quaternet,aksan2019structured,zhang2021we}
or GNNs \cite{jain2016structural,cui2021towards,li2020dynamic,cui2020learning,dang2021msr,mao2019learning,sofianos2021space,ma2022progressively,liu2021motion}
and can be embedded in generative models \cite{gui2018adversarial,hernandez2019human,cui2021efficient,kundu2019bihmp}.
More recent works started to consider relations with surroundings
in the form of intention toward destinations \cite{cao2020long} or
interaction with other entities \cite{corona2020context,adeli2021tripod}.
We advance this line by considering the directional relations between
human and objects beyond the simplistic homogenous graphs such as
in \cite{corona2020context} through a new multi-mechanism adaptive
structural of the persistent-transient duality.

%% file: method.tex
\begin{figure*}[t]
\begin{centering}
\includegraphics[width=0.9\textwidth]{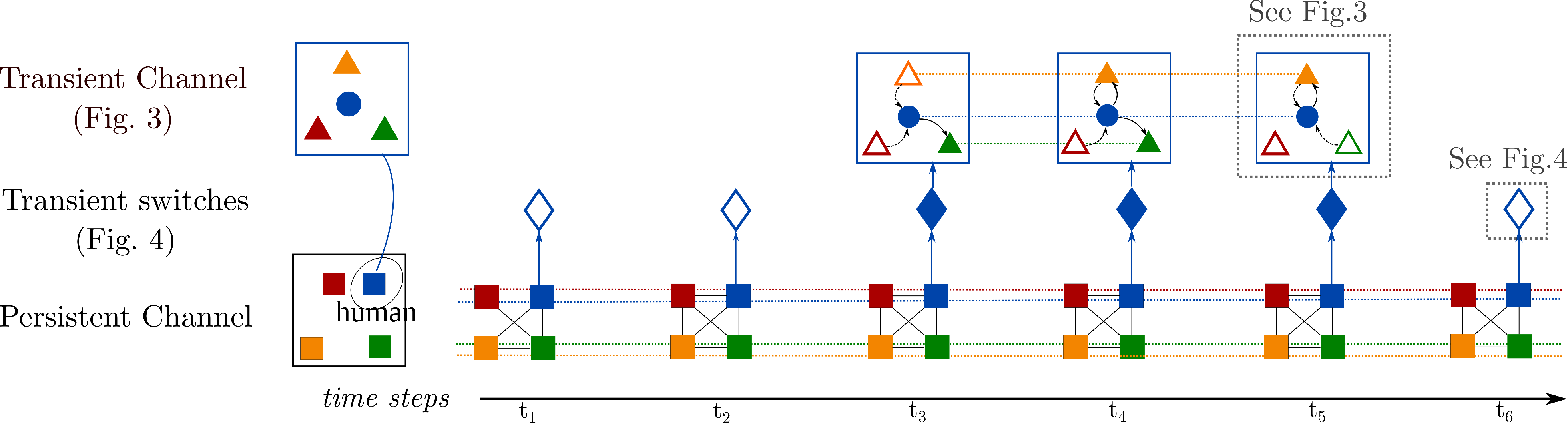}
\par\end{centering}
\caption{The architecture of \emph{Persistent-transient Duality }Networks (PTD).
The \emph{persistent channel }is a heterogeneous graph network connecting
all entities at the same time step. In complement, the \emph{Transient
channel} zooms into the local context of each human (circles) when
they interact with surrounding objects (triangles). This channel works
with the egocentric structure, under the viewpoint of the human subject.
The transient channels are initialized and terminated on-demand, controlled
by the \emph{Transient Switches} (diamonds) which turn on (filled)
or off (empty) the corresponding Transient processes. \label{fig:overal_architecture}}
\end{figure*}

\subsection{Preliminaries}

We consider the problem of modeling the sequential motions of $N$
entities of humans and objects, where each entity $i$ is represented
by a class label $c_{i}$ and sequential visual features $X_{i}=\left\{ x_{i}^{t}\right\} _{t=1}^{T}$.
Given the class labels and the sequential features, we want to predict
the future motions in the next $L$ steps, $Y_{i}=\left\{ y_{i}^{t}\right\} _{t=T+1}^{T+L}$.
Each time instance of features $x_{i}^{t}$ of humans contains skeleton
positions; while those of objects include their object types and bounding
boxes. 

This problem is traditionally approached under the single-mechanism
assumption that the inference dynamics stay the same during the whole
session \cite{corona2020context}. In this section, we present our
new multi-mechanism paradigm, modeled with the persistent-transient
duality, to capture how humans maintain their global operation mode
while swiftly adapt to emerging local interaction situations. 

\subsection{The Persistent-Transient Duality }

We consider the motions of the human bodies and the interacted objects
to be caused by two mechanisms: 1. the navigation according to the
global activity progress; and 2. the individual interactions with
objects, which happens in the local contexts. This multi-mechanism
factorization can be viewed as a persistent-transient process duality:
The \emph{persistent process} operates in large activity temporal
scale and with a global spatial perspective. Its counterpart, the
\emph{transient process,} is local in time and egocentric to each
human subject's spatial viewpoint.

We model this duality by a hierarchical neural network called \emph{Persistent-Transient
Duality Networks (PTD)} whose architecture is drawn in \ref{fig:overal_architecture}.
Within PTD, the \emph{Persistent process} is modeled as a single \emph{Persistent
channel}, which operates recurrently along the whole activity sequence.
Its children, the \emph{Transient channel} instances, are initiated
with personalized structures and representations whenever a human
has a new interaction with the surrounding. When the interaction is
over, the outdated transient channel is terminated, and the control
is returned to the persistent channel waiting in the background. These
transient life cycles are managed by the neural modules called \emph{Transient
Switches}. 

\subsection{Persistent Channel\label{subsec:Persistent-Channel}}

The Persistent channel models the common global view of all humans
and objects in the scene. It takes the form of a heterogeneous graph
attention network \cite{wang2019heterogeneous} with two node types
corresponding to human and object entities and \emph{dense spatial
edges} connecting entities at the same time step. We extend this model
by adding \emph{recurrent temporal edges} along the temporal dimension
for each entity:
\begin{equation}
z_{i}^{t}=\left[x_{i}^{t},m_{i}^{t},m_{i,\mathcal{T}\rightarrow\mathcal{P}}^{t}\right],\hspace{3mm}h_{i}^{t}=\text{RNN}_{c_{i}}\left(z_{i}^{t},h_{i}^{t-1}\right),\label{eq:persistent_rnn}
\end{equation}
where $\text{RNN}_{c_{i}}$ is a recurrent unit (e.g., GRU) that corresponds
to the class of the $i^{\text{th}}$ entity. The input $z_{i}^{t}$
of $\text{RNN}_{c_{i}}$ is formed using the the entity's intrinsic
features $x_{i}^{t}$ (skeleton for humans and bounding boxes for
objects), the spatial messages $m_{i}^{t}$ gathered from spatial
edges, and a message from the active transient channel $m_{i,\mathcal{T}\rightarrow\mathcal{P}}^{t}$
(\ref{subsec:transient_process}). 

The spatial message $m_{i}^{t}$ in \ref{eq:persistent_rnn} is aggregated
from the \emph{spatial edges} of the graph via attention: $m_{i}^{t}=\text{Attn}\left(u_{i}^{t},\left\{ u_{j}^{t}\right\} _{j\neq i}\right)$
with $u\_^{t}=\left[x\_^{t};h\_^{t-1}\right]$. Here and throughout
this work, we use the GAT-based attention function \cite{velivckovic2017graph},
$\text{Attn}\left(q,V\right)$, defined over the query $q$ and the
identical key/value pairs $V=\{v_{j}\}_{j=1}^{N}$ by:\vspace{-2mm}
\begin{equation}
\text{Attn}\left(q,V\right)\coloneqq\sum_{j=1}^{N}\text{softmax}_{j}\left(\sigma\left(\text{MLP}\left(\left[\boldsymbol{W_{q}}q;\boldsymbol{W}_{v}v_{j}\right]\right)\right)\right)\boldsymbol{W}_{v}v_{j},\label{eq:attention_function}
\end{equation}
where $\boldsymbol{W_{q}}$ and $\boldsymbol{W}_{v}$ are the learnable
embedding weights for the query and key/value, $\left[\cdot;\cdot\right]$
is the concatenation, $\sigma\left(\cdot\right)$ is a non-linear
activation.

The Persistent channel generates two outputs from its hidden state:
the message to the transient channel of each human $m_{i,\mathcal{P}\rightarrow\mathcal{T}}^{t}$
and the future position of each entity $\hat{y}_{i,\mathcal{P}}^{t}$:\vspace{-2mm}

\begin{equation}
m_{i,\mathcal{P}\rightarrow\mathcal{T}}^{t}=\text{MLP}\left(h_{i}^{t}\right),\qquad\hat{y}_{i,\mathcal{P}}^{t}=\text{MLP}\left(h_{i}^{t}\right).\label{eq:persistent_output}
\end{equation}
These prediction outputs are later combined with those from the Transient
channel as detailed in \ref{subsec:Future-prediction}.

\subsection{Transient Channel \label{subsec:transient_process}}

Operating with a consistent graph, the Persistent process alone cannot
adapt quickly enough to emerging events that require different perspectives,
particularly when the human interacts with an object. When these cases
are detected, PTD initiates a \emph{Transient process} that (1) zooms
in to the relevant context and (2) take the viewpoint of the subject. 

In our framework, the Transient process is implemented by a neural
\emph{Transient channel}, available one for each \emph{human} entity\emph{.}
We propose an \emph{egocentric recurrent graph networks} for the Transient
channel. The egocentric property of this model is the most important
aspect that separates it from the global view of its parent persistent
channel. This egocentric design reflects in three aspects of \emph{computational
structure, feature representation}, and \emph{inference logic}. 

\paragraph{Egocentric computational structure}

In switching from the global to personalized view, we start by forming
an\emph{ egocentric Transient graph} $\mathcal{G}_{i}^{t}=\left(\mathcal{V}_{i}^{t},\mathcal{E}_{i}^{t}\right)$
at time $t$ of the human $i$ and their relations with the objects
interacted with. For a particular human, subscript $i$ will be omitted
for conciseness. The egocentric characteristic of $\mathcal{G}^{t}$
reflects in its star-like structure: the nodes $\mathcal{V}^{t}$
includes a single \emph{center node $r$} for the considering human,
and \emph{leaf nodes} of indices $\left\{ l\right\} _{l\neq r}$ for
objects. The dynamic edges $\mathcal{E}^{t}$ connect the center with
the leaves in two directions: \emph{inward edges }$e_{l\rightarrow r}^{t}$
reflect which objects the human may consider to interact with, and
the \emph{outward edges} $e_{r\rightarrow l}^{t}$ represents the
objects are being manipulated by the human. In our implementation,
the existences of the edges are determined by thresholding the center-leaf
distances $d_{lr}^{t}$:
\begin{equation}
\begin{cases}
e^{t}\_\in\mathcal{E}^{t} & \text{if }d_{lr}^{t}\leq\beta\_\\
e^{t}\_\notin\mathcal{E}^{t} & \text{otherwise}
\end{cases},\label{eq:ego_graph_structure}
\end{equation}
where $\_$ indicates either the inward or onward, $\beta\_$ is a
pair of threshold hyper-parameters. This thresholding effectively
thins out the neighbors, making the graph more efficient and localized
around the center node. Inward threshold is commonly smaller than
outward one, because humans subject's attention, hence motion, are
affected by objects before and after objects get directly manipulated.
The edges are estimated dynamically for each time step, allowing the
graph's topology to evolve within one single Transient session. 

\begin{figure}
\begin{centering}
\includegraphics[width=0.5\columnwidth]{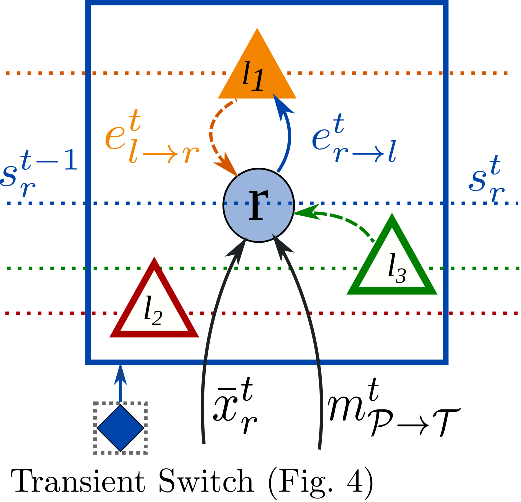}
\par\end{centering}
\caption{The Transient Channel features an egocentric structure of the center
node (circle) and leaves (triangles) with sparse edges (colored arrows).
When being active, it uses the egocentric features $\bar{x}_{r}^{t}$
and the persistent message $m_{\mathcal{P}\rightarrow\mathcal{T}}^{t}$
to update its state $s_{r}^{t}$. The life-cycle of this channel is
determined by the Transient Switch (blue diamond) visualized in \ref{fig:transient-switch}.\label{fig:transient-channel}}
\end{figure}

\paragraph{Egocentric representation.}

\begin{figure}
\begin{centering}
\includegraphics[width=0.47\columnwidth]{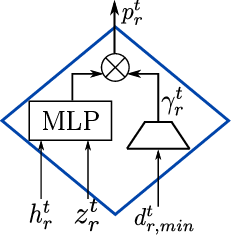}
\par\end{centering}
\caption{The Transient Switch predicts the transient score $p_{r}^{t}$ from
the hidden states $h_{r}^{t}$ and the input $z_{r}^{t}$ of the persistent
RNN. This score is modulated by a discount factor $\gamma_{r}^{t}$
computed from the minimum distance $d_{r,min}^{t}$ via the discount
gate (trapezoid). \label{fig:transient-switch}}
\end{figure}

Paired with the egocentric graph structure, the geometrical features
of the entities are also transformed into the egocentric coordinate
system corresponding to the viewpoint of the human center node $r$:
\begin{align}
\bar{x}\_^{t} & =f_{\text{ego}}\left(x\_^{t},x_{r}^{t}\right)=x\_^{t}-\text{centroid}(x_{r}^{t}),\label{eq:egocentric_leaf}
\end{align}
where $\_$ indicate both center or leaf nodes. Conceptually, this
change of system puts various patterns of the human's motion into
the same aligned space. In the global view, features of two similar
actions can be vastly different if they are far away. After this transformation,
they are aligned into the common egocentric view of the center human
subject. This alignment filters out the irrelevant global information,
and facilitates efficient inference of the egocentric model.

\paragraph{Egocentric inference.}

Operating on the transient graph structure, the RNN hidden state of
each node $s_{-}^{t}$ is updated with the input aggregated through
attention-based message passing along the edges in $\mathcal{E}^{t}$.
See \ref{fig:transient-channel} for an illustration.

In detail, for the \emph{center node}, the inward messages from its
leaves are aggregated into: $m_{r}^{t}=\text{Attn}\left(\begin{array}{l}
u_{r}^{t},\left\{ u_{l}^{t}\right\} _{e_{l\rightarrow r}^{t}\in\mathcal{E}^{t}}\end{array}\right)$, where $u^{t}\_=\left[\bar{x}\_^{t};s\_^{t-1}\right]$ and $\text{Attn}\left(q,V\right)$
is defined in \ref{eq:attention_function}. This message is combined
with the egocentric features $\bar{x}_{r}^{t}$ and the persistent
channel's message $m_{\mathcal{P}\rightarrow\mathcal{T}}^{t}$ (\ref{eq:persistent_output})
to update the recurrent state $s_{r}^{t}$:
\begin{equation}
s_{r}^{t}=\text{RNN}_{r}\left(\left[\bar{x}_{r},m_{r}^{t},m_{\mathcal{P}\rightarrow\mathcal{T}}^{t}\right],s_{r}^{t-1}\right).\label{eq:center_gru}
\end{equation}

For\emph{ leaf nodes}, when they are interacted by the center node
(indicated via $e_{r\rightarrow l}^{t}$), they receive a center-broadcasted
message $m_{l}^{t}=\text{MLP}\left(\left[\bar{x}_{r}^{t};s_{r}^{t-1}\right]\right)$
and update their state $s_{l}^{t}$:

\begin{equation}
s_{l}^{t}=\begin{cases}
\text{\text{RNN}}_{l}\left(\left[\bar{x}_{l}^{t};m_{l}^{t}\right],s_{l}^{t-1}\right) & \textrm{if }e_{r\rightarrow l}^{t}\in\mathcal{E}^{t}\\
s_{l}^{t-1} & \text{otherwise}
\end{cases},\label{eq:local_leaf_hidden_states}
\end{equation}

The updated hidden states are used to generate the messages sent to
the persistent channel $m_{\mathcal{T}\rightarrow\mathcal{P}}^{t}$
and the transient predictions about the future position $\hat{y}_{-}^{\mathcal{T},t}$
of the entities:\vspace{-2mm}

\begin{align}
m_{\mathcal{T}\rightarrow\mathcal{P}}^{t}=\text{MLP}\left(s_{r}^{t}\right),\quad\hat{y}_{-,\mathcal{T}}^{t} & =f_{\text{ego}}^{-1}\left(\text{MLP}\left(s_{-}^{t}\right)\right),\label{eq:transient-readout}
\end{align}
where $f_{\text{ego}}^{-1}$ is the inverse function of \ref{eq:egocentric_leaf}
converting the egocentric back to global coordinates. The Transient
predictions $\hat{y}_{-}^{\mathcal{T},t}$ are then combined with
the Persistent counterparts as described in \ref{subsec:Future-prediction}.
As a key modeling feature, the egocentric design plays a crucial goal
in the power of the PTD which will be demonstrated later in the Ablation
studies (\ref{subsec:HOI-Ablation}).

\subsection{Switching Transient Processes \label{subsec:switch}}

The life cycles of the Transient processes are managed based on the
situation of human's activity. These decisions are made by a neural
\emph{Transient Switch} (See \ref{fig:transient-switch}) that makes
switch-on and switch-off decisions of the Transient process for each
human entity. The switch-on probability $p_{r}^{t}$ is computed by
considering the current hidden state $h_{r}^{t}$ and the input $z_{r}^{t}$
of the persistent RNN in \ref{eq:persistent_rnn}:

\begin{equation}
\hat{p}_{r}^{t}=\gamma_{r}^{t}\cdot\text{sigmoid}\left(\text{MLP}\left(\left[h_{r}^{t};z_{r}^{t}\right]\right)\right),\label{eq:switch_score}
\end{equation}
where $h_{r}^{t}$ and $z_{r}^{t}$ is the current hidden state and
the input of the persistent RNN. The discount factor $\gamma_{r}^{t}\in[0,1]$
responds to subject's distance to the nearest object $d_{r,min}^{t}$:
\begin{equation}
\gamma_{r}^{t}=\exp\left(-\eta\cdot d_{r,min}^{t}\right),\label{eq:discount_factor}
\end{equation}
where $\eta$ is a learnable decay rate. This factor acts as a disruptive
shortcut gate that modulates the switching decision based on the spatial
evidence of the interaction. 

Finally, the binary switch decision is made by thresholding the score
with a learnable threshold $\theta$: the switch is $on$ when $\hat{p}_{r}^{t}>=\theta$,
and is \emph{off} otherwise. When the switch changes from \emph{off}
to \emph{on}, a new Transient process is created for the subject.
It will run until the switch turns \emph{off}, then the persistent
process again becomes the single operator.

\subsection{Future prediction\label{subsec:Future-prediction}}

In PTD, future motions are predicted autoregressively: After running
through the observed sequence $X^{1:T}$, PTD keeps unrolling to predict
the requested $L$ time steps and feeds its prediction back as input
to keep unrolling. 

At each future time step $t$, the predictions from Persistent and
Transient channels $\hat{Y}_{\mathcal{P}}^{t}$, $\hat{Y}_{\mathcal{T}}^{t}$
(\ref{eq:persistent_output}, \ref{eq:transient-readout}) are combined
with the priority on the Transient predictions. For a human entity,
if its Transient channel is activated, the Transient prediction will
be chosen; otherwise, the Persistent prediction will be used. For
an object entity, if it receives an active outward Transient edge,
it will take that channel's prediction. If it receives multiple outward
edges, it uses the prediction from the channel with the highest transient
score $\hat{p}_{r}^{t}$. Otherwise, it uses the persistent prediction
by default.

\subsection{Model Training \label{subsec:losses}}

The model is trained end-to-end with three losses: prediction loss
of humans and objects and switch loss:\vspace{-2mm}

\begin{align}
\mathcal{L} & =\lambda_{h}\mathcal{L}_{pred,h}+\lambda_{o}\mathcal{L}_{pred,o}+\lambda_{switch}\mathcal{L}_{switch}.\label{eq:losses}
\end{align}

The \textbf{Prediction loss} $\mathcal{L}_{\text{pred}}$ measures
the mismatch between predicted values $\hat{Y}$ and ground truth
$Y$, implemented as their Euclidean distance:

\begin{equation}
\mathcal{L}_{\text{pred}}=\left\Vert \hat{Y}^{T+1:T+L}-Y^{T+1:T+L}\right\Vert _{2}.\label{eq:pred_loss}
\end{equation}
Even though the Transient switch can be implicitly trained with the
prediction loss, the gradient flowing through this binary gate can
be weak. To directly supervise the Transient switch, we further introduce
the \textbf{Switch loss}: 

\begin{equation}
\mathcal{L}_{switch}=\text{BCE}\left(\hat{P}^{1:T+L},P^{1:T+L}\right),\label{eq:switch_loss}
\end{equation}
where $\hat{P}^{t}$ and $P^{t}$ are the predicted and ground truth
switch scores (\ref{eq:switch_score}) of all human entities at time
step $t$. 

\paragraph{Setting switch ground truth label}

$P^{t}$ from data is an interesting modeling topic. The term represents
the true moment where the human $r$ turns on their ``Transient mode''.
A simple way to learn it is through self-supervision using a binary
label $q^{t}$ on whether an interaction happens at that time. In
particular, it is determined by the outward edges $e_{r\rightarrow l}^{t}$
in the Transient graph (see \ref{subsec:transient_process}): $q^{t}=1$
if $\exists l:e_{r\rightarrow l}^{t}\in\mathcal{E}^{t}$, and $q^{t}=0$
otherwise.

However, is this the true ground truth to human behavior switch? Humans
usually foresee their interaction and change their behavior before
the observable interaction occurs; therefore, $q^{t}$ is actually
too late to turn on the Transient channel. We resolve this mismatch
by using the future ground truth labels of deviations for the current
label of Transient switch: 

\begin{equation}
p^{t}=q^{t}\vee q^{t+1}\vee...\vee q^{t+\omega},\label{eq:groundtruth_switch_label}
\end{equation}
where $\vee$ indicates 'bit-wise or' operator, and $\omega$ is the
postdating window meta-parameter, whose values are examined in the
Supp, Sec. 5.

\paragraph{Multistage training}

We train our model in two stages: we first use teacher-forcing \cite{lecun2015deep}
to use the ground truth position as the input in the prediction stage,
preventing the model from accumulating errors and facilitate faster
training. Then, in the second stage, we fine-tune the model with the
unrolling mechanism introduced in \ref{subsec:Future-prediction}. 

This way of training can be thought of as a curriculum learning technique
\cite{bengio2009curriculum} used in previous works \cite{adeli2021tripod,jain2016structural}
where we initially train the model with an easy problem, then increase
the difficulty of the problem in the later epochs. 

\subsection{Modeling scope and limitations}

The modeling of the PTD in this paper makes an assumption that the
Transient processes of different persons are independent of each other.
Extensions can be made to break this assumption and support more complex
cases, such as collaborative and competitive multi-agent systems. 

The model also assumes that there is a hard temporal border between
the persistent and transient processes, which may be an approximation
as humans sometimes have a mixed-up thinking and acting mechanisms.
Future works may involve allowing the two processes to take over each
other more softly, hence can cover these cases.

The PTD formulation in this section is done particularly for HOI-M
forecasting. Modifications may be necessary for other applications.
Example adaptation for \emph{pedestrian trajectory prediction} is
given in Supp. Sec 9.

%% file: exp.tex
We examine the effectiveness of PTD via quantitative evaluations,
generalization trial, visual analysis, and ablation studies on two
HOI-M prediction datasets: WBHM \cite{Mandery2015a} and Bimanual
Actions \cite{dreher2020learning}. 

\subsection{Preliminaries}

This section describes the datasets, the baselines, and metrics in
used. Further details are in Supp, Sec. 1.

\subsubsection{Datasets}

\paragraph{Whole-Body Human Motion (WBHM) Database \cite{Mandery2015a}}

is a large-scale dataset featuring 3D motion data of both humans and
objects, which is well suited for this paper. From the raw 3D data,
the selected visual features include 3D skeleton poses of 18 joints
for human entities $(x\in\mathbb{R}^{54})$ and 3D bounding boxes
for objects $(x\in\mathbb{R}^{24})$, sampled at 10Hz, consistent
with the compared methods \cite{corona2020context}.

\paragraph{Bimanual Actions Dataset \cite{dreher2020learning} }

contains activities of subjects using both hands to interact with
different objects at the same time. Unlike WBHM with 3D geometrical
features, all features here are in 2D coordinates making the dataset
more challenging for motion prediction. Furthermore, the subjects
in this dataset always use two arms to interact concurrently with
different objects, requiring a new capability of modeling the collaboration
between the arms. PTD naturally support this use case by considering
each arm to be one human entity, each with features of 2D locations
of the arm key points and hand bounding box.

\subsubsection{Compared methods and baselines}

We compare PTD with the \emph{HOI-M forecasting SOTAs}: CRNN-OPM,
CRNN-OPM-LI \cite{corona2020context} and the\emph{ pose forecasting
SOTAs}: STS-GCN \cite{sofianos2021space}, Motion-Mixer \cite{bouazizi2022motionmixer}.
Motion-Mixer and STS-GCN are retrained using the provided codes\footnote{https://github.com/FraLuca/STSGCN

https://github.com/MotionMLP/MotionMixer}. The others are re-implemented with the settings provided in the
original papers. We also use several common baseline methods of Zero
Velocity, Running avg. 2, GRU \cite{martinez2017human}. 

\subsubsection{Evaluation metric}

The prediction errors at each time step are calculated as the Euclidean
distances with the ground-truth for both human poses and object bounding
boxes (mm for WBHM, pixel for Bimanual Action Dataset). The error
for a sequence is then computed as the average errors across $L$
prediction frames. Prediction performances are reported as the mean
and std of the average errors of 5 independent runs for humans and
objects entities. 

\subsection{Motion forecasting on WBHM Dataset \label{subsec:exp_wbhm}}

For WBHM dataset, we follow the common evaluation protocol \cite{corona2020context}:
to observe for 1 second ($T=10$) and predict the next 2 seconds ($L=20$). 

\paragraph{Quantitative evaluation.}

The average errors reported in \ref{tab:quantitative_wbhm} clearly
indicates that PTD consistently outperforms the SoTAs in both human
pose and object box forecasting. Supp, Sec.3 further reports detailed
errors at each time step.

\begin{table}
\begin{centering}
\begin{tabular}{lcc}
\toprule 
\multirow{1}{*}{} & {\footnotesize{}Human} & {\footnotesize{}Obj}\tabularnewline
\midrule
{\footnotesize{}Zero-Velocity} & {\footnotesize{}176.45} & {\footnotesize{}128.6}\tabularnewline
{\footnotesize{}Running avg. 2} & {\footnotesize{}183.95} & {\footnotesize{}133.3}\tabularnewline
{\footnotesize{}GRU \cite{martinez2017human}} & {\footnotesize{}102.86 $\pm$ 1.4} & {\footnotesize{}119.64 $\pm$ 1.6}\tabularnewline
{\footnotesize{}STS-GCN \cite{sofianos2021space}} & {\footnotesize{}101.36 $\pm$ 2.4} & {\footnotesize{}-}\tabularnewline
{\footnotesize{}Motion-Mixer \cite{bouazizi2022motionmixer}} & {\footnotesize{}87.35 $\pm$ 1.2} & {\footnotesize{}-}\tabularnewline
{\footnotesize{}CRNN-OPM \cite{corona2020context}} & {\footnotesize{}99.01 $\pm$ 1.1} & {\footnotesize{}87.52 $\pm$ 1.6}\tabularnewline
{\footnotesize{}CRNN-OPM-LI \cite{corona2020context}} & {\footnotesize{}95.96 $\pm$ 1.7} & {\footnotesize{}74.27 $\pm$ 1.3}\tabularnewline
\midrule 
\textbf{\footnotesize{}PTD (Ours)} & \textbf{\footnotesize{}85.53}{\footnotesize{} $\pm$}\textbf{\footnotesize{}
0.9} & \textbf{\footnotesize{}70.69}{\footnotesize{} $\pm$}\textbf{\footnotesize{}
0.5}\tabularnewline
\bottomrule
\end{tabular}
\par\end{centering}
\caption{The average errors (mm) on WBHM after 5 independent runs. PTD outperforms
other SOTAs in both human and object prediction. The errors at each
time step are given in Supp, Sec. 3.\label{tab:quantitative_wbhm}}
\end{table}

\paragraph{Visual analysis.}

\begin{figure}[t]
\begin{centering}
\includegraphics[width=0.98\columnwidth]{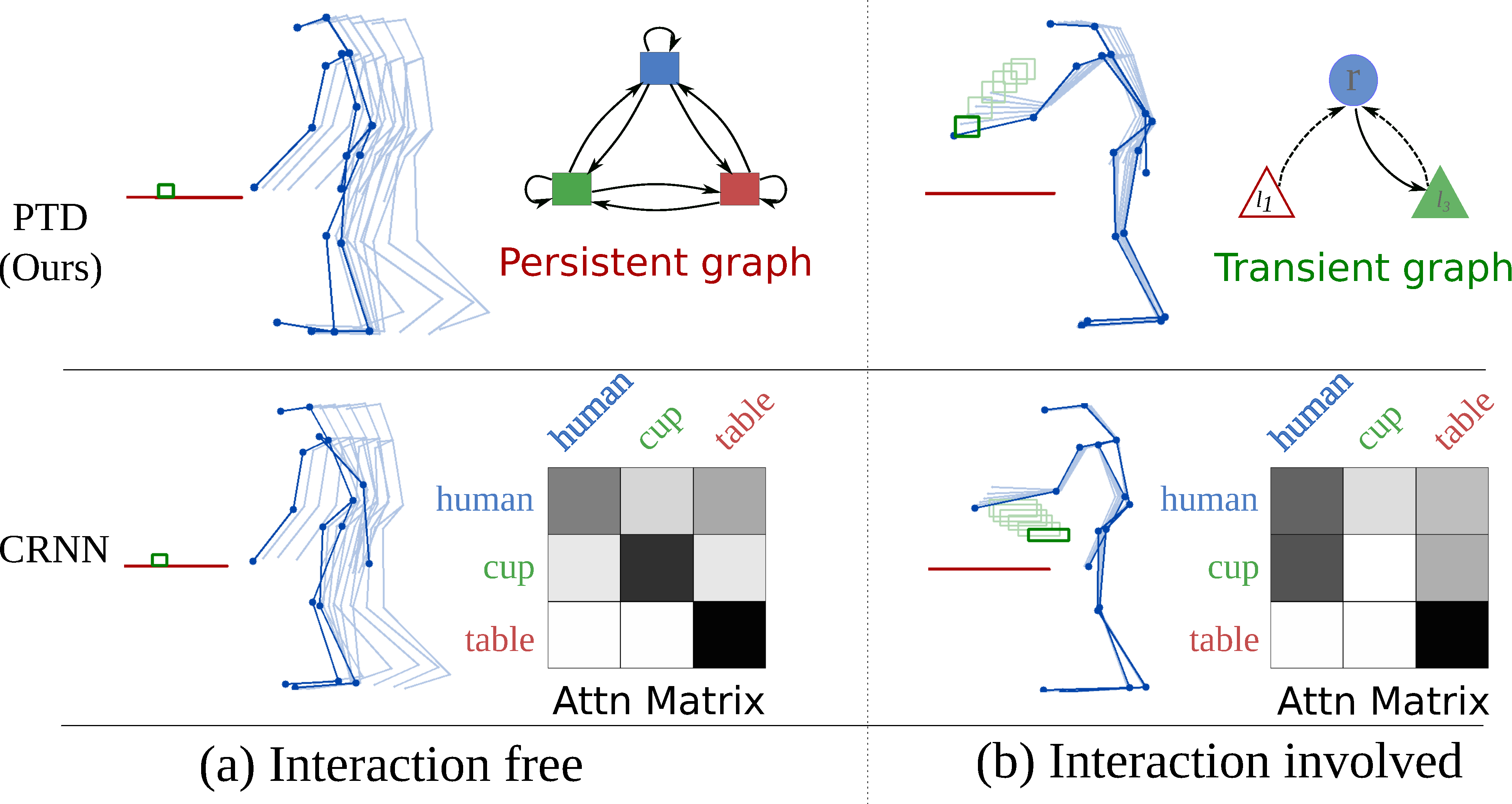}
\par\end{centering}
\caption{Visual Comparison in WBHM. When the situation changes from interaction-free
(a) to interaction-involved (b), PTD \emph{(Upper row}) switches on
its Transient channel with egocentric structures and handles the interaction
accurately; In contrast, CRNN-OPM-LI \cite{corona2020context} (\emph{Lower
row}) uses a single mechanism, resulting in the sluggish adaptation
of the attention map and inaccurate predictions. \label{fig:qualitative_hoi}}
\end{figure}

\begin{figure}
\centering{}\includegraphics[width=0.98\columnwidth]{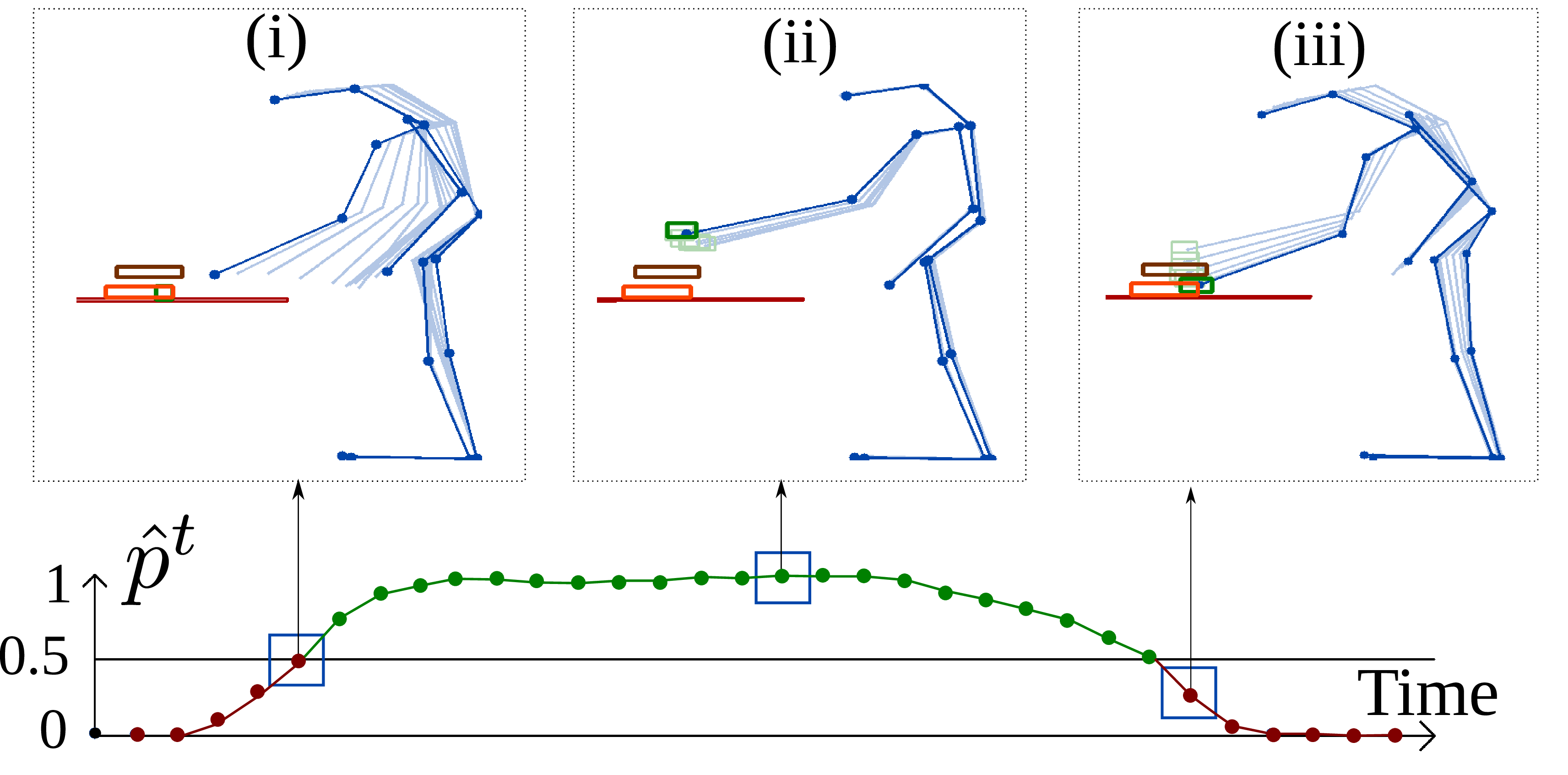}\caption{The switching behavior in WBHM. The Transient Switch anticipates the
beginning (i), stays stable during (ii), and anticipates the end (iii)
of the interaction. This early anticipation is crucial for timely
process switching. \label{fig:switch_hoi_wbhm}}
\end{figure}

We verify the benefit of our model by visualizing the internal output
predictions and graph structures of PTD compared to CRNN-OPM-LI \cite{corona2020context}.
The upper row of \ref{fig:qualitative_hoi} shows that PTD could learn
to switch the mechanism from Persistent to Transient when the situation
changes from interaction-free (a) to interaction-involved (b). The
Transient graph in \emph{(b)} reflects the interactions correctly
thanks to it being trained on targeted samples.

In contrast, CRNN-OPM-LI \cite{corona2020context} (lower row) holds
on to a single global mechanism and does not evolve adequately for
the swift change in the true relational topology, resulting in inaccurate
and unrealistic interactions.

The operation of the Transient Switch is visualized in \ref{fig:switch_hoi_wbhm}.
When the interaction is about to occur (i), the switch score $\hat{p}_{r}^{t}$
(\ref{eq:switch_score}) increases to reflect the prospective of the
interaction. When the score passes the threshold, it switches on the
Transient channel a few time-steps before the interaction can be observed,
precisely as designed (\ref{eq:groundtruth_switch_label}). After
maintaining high values during interaction (ii), the score falls when
it anticipates the end of the interaction, deactivating the transient
channel (iii).

\paragraph{Generalization analysis.}

\begin{figure}
\begin{minipage}[t]{0.49\columnwidth}%
\includegraphics[width=1\columnwidth]{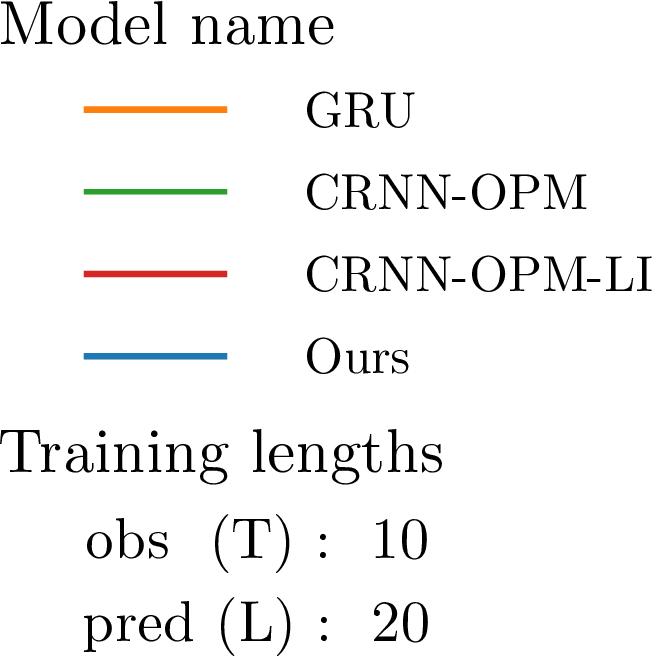}%
\end{minipage}\hfill{}\subfloat[Keep $L=20$, vary $T$\label{fig:vary_obs}]{\includegraphics[width=0.49\columnwidth]{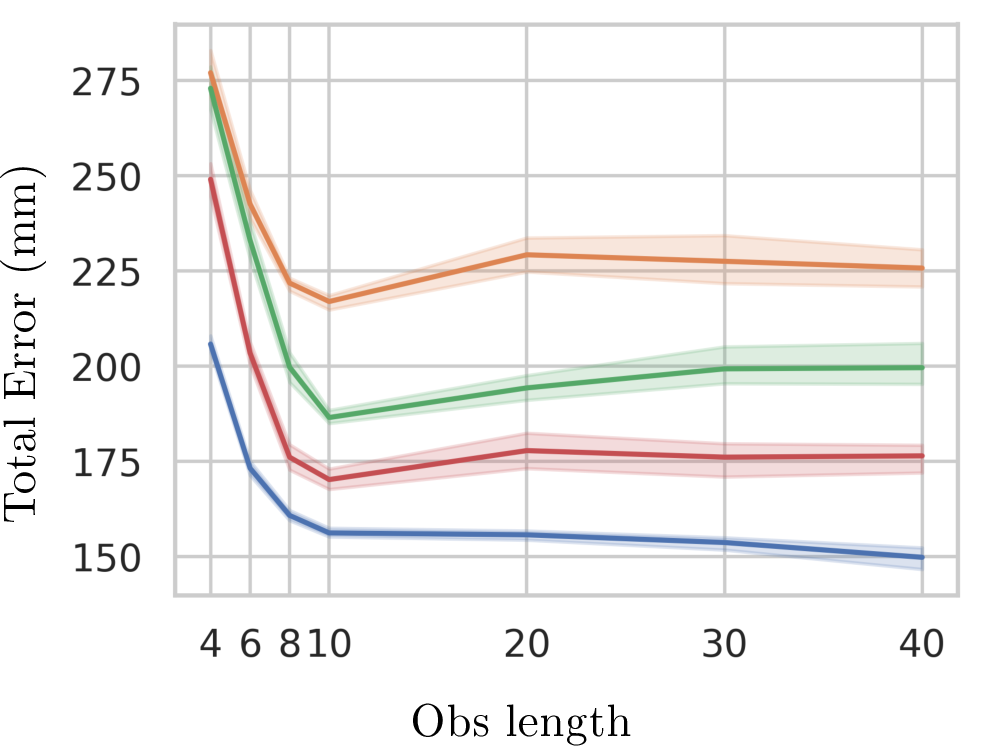}}\hfill{}\subfloat[Keep $T=10$, vary $L$\label{fig:vary_pred}]{\includegraphics[width=0.49\columnwidth]{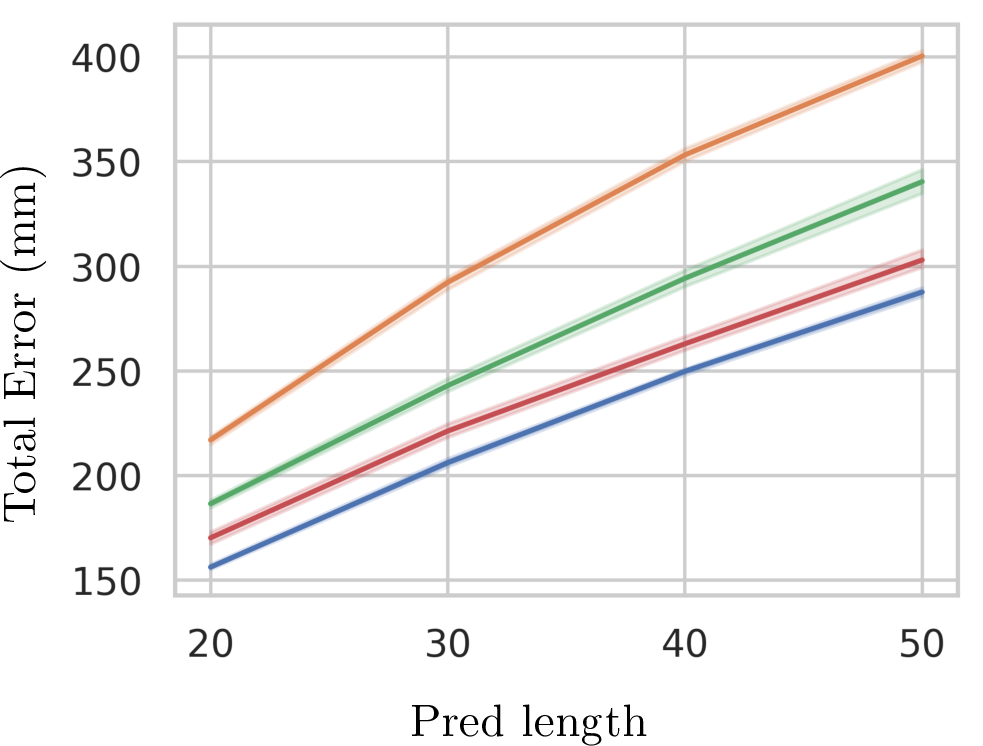}}\hfill{}\subfloat[Increase $T=30$, vary $L$ \label{fig:vary_both}]{\includegraphics[width=0.49\columnwidth]{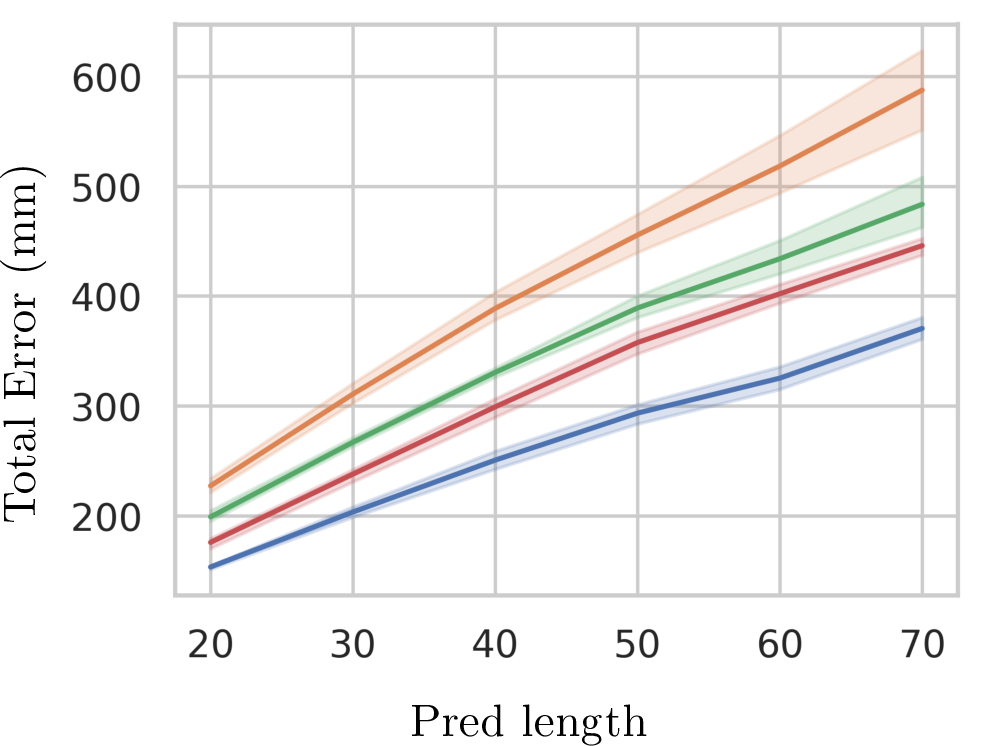}}

\caption{Generalizability test on varying observed lengths ($T$) and predicted
lengths ($L)$ of the sequences in testing from the ones used in training.\label{fig:generalization_analysis}}
\end{figure}

With more accurate modeling, PTD promises a greater generalization
in predicting the patterns unseen in training. To evaluate such potential,
we use a generalization test \cite{hupkes2020compositionality} to
set up a series of experiments where PTD and other models compete
on the test sequences of lengths different from the ones used in training:
(1) observation length $T$ varies, (2) prediction length $L$ varies,
and (3) both observation and prediction lengths varies. 

The results are measured as the total average errors (mm) of humans
and objects and are visualized in \ref{fig:generalization_analysis}.
We exclude STS-GCN and MotionMixer due to their dependence on the
sequence length. 

\textbf{1. Varying observation length $T$} (\ref{fig:vary_obs}):
Interestingly, when observing longer sequences in testing ($T>10$),
the baseline models failed to generalize and perform worse. In contrast,
with mechanism switching, PTD is flexible enough to take advantage
of the longer observed data to gain performance without re-training.
Also, with shorter observed sequences ($T<10$), PTD is more resilient
than other models, showing its capacity of modeling the generic pattern
and avoid overfitting. 

\textbf{2. Predicting longer sequences. }With same observation $(T=10)$
and increased prediction length $L$ of the testing data (\ref{fig:vary_pred}),
PTD keeps the superior performance by extrapolating well to farther
future predictions.

\textbf{3. Increasing both lengths. }Finally, when both length changes
(\ref{fig:vary_both}), the result is consistent with case 1. and
2. where PTD shows its superior ability to take advantage of more
data and generalize well to longer predictions.

\paragraph{Ablation studies.\label{subsec:HOI-Ablation}}

\begin{table}
\centering{}%
\begin{tabular*}{1\columnwidth}{@{\extracolsep{\fill}}l>{\raggedright}m{0.4\columnwidth}>{\centering}m{0.15\columnwidth}>{\centering}m{0.15\columnwidth}}
\toprule 
\multirow{1}{*}{} & \multirow{1}{0.4\columnwidth}{\centering{}{\footnotesize{}Ablation}} & {\footnotesize{}Human} & {\footnotesize{}Object}\tabularnewline
\midrule
{\footnotesize{}1} & {\footnotesize{}w/o Transient channel} & {\footnotesize{}89.49} & {\footnotesize{}74.91}\tabularnewline
{\footnotesize{}2} & {\footnotesize{}w/o Persistent channel} & {\footnotesize{}91.41} & {\footnotesize{}78.67}\tabularnewline
{\footnotesize{}3} & {\footnotesize{}w/o egocentric property} & {\footnotesize{}87.90} & {\footnotesize{}72.30}\tabularnewline
{\footnotesize{}4} & {\footnotesize{}w/o heuristic switch} & {\footnotesize{}86.44} & {\footnotesize{}74.85}\tabularnewline
{\footnotesize{}5} & {\footnotesize{}w/o $\gamma$} & {\footnotesize{}86.17} & {\footnotesize{}70.71}\tabularnewline
{\footnotesize{}6} & {\footnotesize{}w/ only $\gamma$} & {\footnotesize{}87.11} & {\footnotesize{}76.95}\tabularnewline
{\footnotesize{}7} & {\footnotesize{}w/o switching transient} & {\footnotesize{}86.00} & {\footnotesize{}71.40}\tabularnewline
{\footnotesize{}8} & {\footnotesize{}w/o switch loss} & {\footnotesize{}89.59} & {\footnotesize{}73.17}\tabularnewline
{\footnotesize{}9} & {\footnotesize{}w/o multistage training} & {\footnotesize{}90.21} & {\footnotesize{}72.22}\tabularnewline
\midrule 
 & \textbf{\footnotesize{}Full PTD model} & \textbf{\footnotesize{}85.53} & \textbf{\footnotesize{}70.69}\tabularnewline
\bottomrule
\end{tabular*}\caption{Ablation studies on WBHM (avg. errors in mm.) \label{tab:ablation-results}}
\end{table}

We examine the roles of PTD's core components by making ablations
from the model and report the performances in \ref{tab:ablation-results}.
They include:

1.\textbf{ Without Transient channel}: Being alone, Persistent Channel
performs significantly worse than when with its Transient partner
in the duality.

2. \textbf{Without Persistent channel: }Transient Channel also struggles
and gives worse performance operating alone. 

3. \textbf{Without Egocentric property}, including egocentric structure
(\ref{eq:ego_graph_structure}) and representation (\ref{eq:egocentric_leaf}),
Transient channels operate on a fully-connected graph with global
features. This ablated model suffers a significant drop in performance,
showing the role of egocentric design.

4.\textbf{ Heuristic switch:} This experiment probe the need for the
\emph{Transient Switch} (\ref{subsec:switch}) by replacing it with
a heuristic rule $\hat{p}_{r}^{t}=d_{r,min}^{t}\leq\beta$. This heuristic
switch can still provide benefit compared to single channels (row
1,2). However, being too stiff, it cannot represent the switching
patterns and fails to reach the duality's full potential. 

5. \textbf{Switch without spatial discount factor: }Without $\gamma_{r}^{t}$
(\ref{eq:switch_score}), the \emph{Switch} could not respond fast
enough to context changes, resulting in slightly weaker performance.

6. \textbf{Switch with only discount factor: }This quick-change factor
could not do the job by itself as it is susceptible to noisy patterns,
performing even worse than row 5. 

7. \textbf{Without switching transient:} The transient channels is
always on and may capture irrelevant patterns outside the HOI-context,
leading to a drop in the performance.

8. \textbf{Without switch loss: }We study the role of the switch's
direct supervision (\ref{subsec:losses}) by setting $\lambda_{switch}$
to 0, taking $\mathcal{L}_{switch}$ out of \ref{eq:losses}. This
unsupervised switch only relies on weak gradient flowing back from
prediction loss and delivers significantly weakened performance. 

9. \textbf{Without multistage training: }Without such a training procedure
(\ref{subsec:losses}), the model suffers from accumulating losses
during early epochs and capture less accurate motion pattern, resulting
in a decrease in the performance.

\subsection{Motion forecasting on Bimanual Action Dataset}

While WBHM is rich in patterns and is well-suited for large scale
analysis, the Bimanual Action is more realistic, as its 2D RGB data
are more available. Due to the increased ambiguity of the 2D data,
we use longer observed (2s/20 steps) and shorter predict lengths (1s/10
steps).

\textbf{Quantitative evaluation:} The average errors measured in pixel
are reported in \ref{tab:quantitative_bimanual}. Consistent with
the results in WBHM dataset, PTD outperforms other SOTAs in both human
and object motion forecasting. The errors at each time steps are also
detailed in Supp, Sec. 3. \textbf{Visual analysis} of PTD's operation
in this dataset (similar to that of WBHM in \ref{fig:qualitative_hoi}
and \ref{fig:switch_hoi_wbhm}) is provided in Supp, Sec 4. 

\begin{table}
\begin{centering}
\begin{tabular}{lccc}
\toprule 
\multirow{1}{*}{} & {\footnotesize{}Arm Keypoints} & {\footnotesize{}Hand} & {\footnotesize{}BoxObj}\tabularnewline
\midrule 
{\footnotesize{}Zero-Velocity} & {\footnotesize{}12.11} & {\footnotesize{}21.52} & {\footnotesize{}7.02}\tabularnewline
{\footnotesize{}Running avg. 2} & {\footnotesize{}12.80} & {\footnotesize{}22.19} & {\footnotesize{}7.30}\tabularnewline
{\footnotesize{}GRU \cite{martinez2017human}} & {\footnotesize{}12.37 $\pm$ 0.4} & {\footnotesize{}20.80 $\pm$ 0.9} & {\footnotesize{}7.04 $\pm$ 0.0}\tabularnewline
{\footnotesize{}STS-GCN \cite{sofianos2021space}} & {\footnotesize{}11.85 $\pm$ 0.5} & {\footnotesize{}-} & {\footnotesize{}-}\tabularnewline
{\footnotesize{}Motion-Mixer \cite{bouazizi2022motionmixer}} & {\footnotesize{}11.68 $\pm$ 0.2} & {\footnotesize{}-} & {\footnotesize{}-}\tabularnewline
{\footnotesize{}CRNN-OPM \cite{corona2020context}} & {\footnotesize{}12.81 $\pm$ 0.1} & {\footnotesize{}21.66 $\pm$ 0.2} & {\footnotesize{}7.16 $\pm$ 0.1}\tabularnewline
{\footnotesize{}CRNN-OPM-LI \cite{corona2020context}} & {\footnotesize{}11.97 $\pm$ 0.3} & {\footnotesize{}20.13 $\pm$ 0.2} & {\footnotesize{}7.05 $\pm$ 0.1}\tabularnewline
\midrule 
\textbf{\footnotesize{}PTD (Ours)} & \textbf{\footnotesize{}10.94}{\footnotesize{} $\pm$ }\textbf{\footnotesize{}0.2} & \textbf{\footnotesize{}18.80}{\footnotesize{} $\pm$ }\textbf{\footnotesize{}0.1} & \textbf{\footnotesize{}6.81}{\footnotesize{} $\pm$ }\textbf{\footnotesize{}0.0}\tabularnewline
\bottomrule
\end{tabular}
\par\end{centering}
\caption{The avg. errors (pixel) on Bimanual Action dataset from five different
runs.\label{tab:quantitative_bimanual}}
\end{table}

\subsection{Empirical Complexity Analysis}

We did an empirical analysis on the model size and observed that \emph{PTD
has a comparable number of parameters to other methods}. Detailed
numeric analysis is shown in Supp, Sec 2. This fact confirms that
the good performance of PTD is caused by the new multi-mechanism scheme
without the negative trade-off in computation cost.

%% file: discuss.tex
In this work, we have introduced a new concept of Persistent-Transient
Duality to model the multi-mechanism nature of human behavior in interaction
with objects. The duality is implemented into a parent-children network
that demonstrates its effectiveness and generalizability through extensive
evaluations in HOI-M forecasting. 

Given the ubiquity of persistent-transient relationship in human behavior
and the genericity of PTD design, the model can be readily extended
to other applications including \emph{pedestrian trajectory prediction}
(as demonstrated in Supp., Sec 9) and potentially \emph{social interaction
modeling}, and \emph{human-machine collaboration}. Future development
may include more fluid dynamics between the two processes and allowing
multiple transient instances.

%% file: main.bbl
\begin{thebibliography}{10}\itemsep=-1pt

\bibitem{adeli2021tripod}
Vida Adeli, Mahsa Ehsanpour, Ian Reid, Juan~Carlos Niebles, Silvio Savarese,
  Ehsan Adeli, and Hamid Rezatofighi.
\newblock Tripod: Human trajectory and pose dynamics forecasting in the wild.
\newblock In {\em Proceedings of the IEEE/CVF International Conference on
  Computer Vision}, pages 13390--13400, 2021.

\bibitem{ajzen1985intentions}
Icek Ajzen.
\newblock {\em From intentions to actions: A theory of planned behavior}.
\newblock Springer, 1985.

\bibitem{aksan2019structured}
Emre Aksan, Manuel Kaufmann, and Otmar Hilliges.
\newblock Structured prediction helps 3d human motion modelling.
\newblock In {\em Proceedings of the IEEE/CVF International Conference on
  Computer Vision}, pages 7144--7153, 2019.

\bibitem{baker2014fast}
Adam~P Baker, Matthew~J Brookes, Iead~A Rezek, Stephen~M Smith, Timothy
  Behrens, Penny J~Probert Smith, and Mark Woolrich.
\newblock Fast transient networks in spontaneous human brain activity.
\newblock {\em Elife}, 3:e01867, 2014.

\bibitem{bengio2009curriculum}
Yoshua Bengio, J{\'e}r{\^o}me Louradour, Ronan Collobert, and Jason Weston.
\newblock Curriculum learning.
\newblock In {\em Proceedings of the 26th annual international conference on
  machine learning}, pages 41--48, 2009.

\bibitem{bouazizi2022motionmixer}
Arij Bouazizi, Adrian Holzbock, Ulrich Kressel, Klaus Dietmayer, and Vasileios
  Belagiannis.
\newblock Motionmixer: Mlp-based 3d human body pose forecasting.
\newblock {\em arXiv preprint arXiv:2207.00499}, 2022.

\bibitem{cao2020long}
Zhe Cao, Hang Gao, Karttikeya Mangalam, Qi-Zhi Cai, Minh Vo, and Jitendra
  Malik.
\newblock Long-term human motion prediction with scene context.
\newblock In {\em European Conference on Computer Vision}, pages 387--404.
  Springer, 2020.

\bibitem{corona2020context}
Enric Corona, Albert Pumarola, Guillem Alenya, and Francesc Moreno-Noguer.
\newblock Context-aware human motion prediction.
\newblock In {\em Proceedings of the IEEE/CVF Conference on Computer Vision and
  Pattern Recognition}, pages 6992--7001, 2020.

\bibitem{cui2021towards}
Qiongjie Cui and Huaijiang Sun.
\newblock Towards accurate 3d human motion prediction from incomplete
  observations.
\newblock In {\em Proceedings of the IEEE/CVF Conference on Computer Vision and
  Pattern Recognition}, pages 4801--4810, 2021.

\bibitem{cui2021efficient}
Qiongjie Cui, Huaijiang Sun, Yue Kong, Xiaoqian Zhang, and Yanmeng Li.
\newblock Efficient human motion prediction using temporal convolutional
  generative adversarial network.
\newblock {\em Information Sciences}, 545:427--447, 2021.

\bibitem{cui2020learning}
Qiongjie Cui, Huaijiang Sun, and Fei Yang.
\newblock Learning dynamic relationships for 3d human motion prediction.
\newblock In {\em Proceedings of the IEEE/CVF Conference on Computer Vision and
  Pattern Recognition}, pages 6519--6527, 2020.

\bibitem{dang2021msr}
Lingwei Dang, Yongwei Nie, Chengjiang Long, Qing Zhang, and Guiqing Li.
\newblock Msr-gcn: Multi-scale residual graph convolution networks for human
  motion prediction.
\newblock In {\em Proceedings of the IEEE/CVF International Conference on
  Computer Vision}, pages 11467--11476, 2021.

\bibitem{dreher2020learning}
Christian R.~G. Dreher, Mirko Wächter, and Tamim Asfour.
\newblock Learning object-action relations from bimanual human demonstration
  using graph networks.
\newblock {\em IEEE Robotics and Automation Letters (RA-L)}, 5(1):187--194,
  2020.

\bibitem{duncan2019chemical}
T~Michael Duncan and Jeffrey~A Reimer.
\newblock {\em Chemical Engineering Design and Analysis}.
\newblock Cambridge University Press, 2019.

\bibitem{escorcia2013spatio}
Victor Escorcia and Juan Niebles.
\newblock Spatio-temporal human-object interactions for action recognition in
  videos.
\newblock In {\em Proceedings of the IEEE International Conference on Computer
  Vision Workshops}, pages 508--514, 2013.

\bibitem{fragkiadaki2015recurrent}
Katerina Fragkiadaki, Sergey Levine, Panna Felsen, and Jitendra Malik.
\newblock Recurrent network models for human dynamics.
\newblock In {\em Proceedings of the IEEE International Conference on Computer
  Vision}, pages 4346--4354, 2015.

\bibitem{ghosh2017learning}
Partha Ghosh, Jie Song, Emre Aksan, and Otmar Hilliges.
\newblock Learning human motion models for long-term predictions.
\newblock In {\em 2017 International Conference on 3D Vision (3DV)}, pages
  458--466. IEEE, 2017.

\bibitem{ghosh2020stacked}
Pallabi Ghosh, Yi Yao, Larry Davis, and Ajay Divakaran.
\newblock Stacked spatio-temporal graph convolutional networks for action
  segmentation.
\newblock In {\em Proceedings of the IEEE/CVF Winter Conference on Applications
  of Computer Vision}, pages 576--585, 2020.

\bibitem{greenwood1991electrical}
Allan Greenwood.
\newblock Electrical transients in power systems.
\newblock 1991.

\bibitem{gui2018adversarial}
Liang-Yan Gui, Yu-Xiong Wang, Xiaodan Liang, and Jos{\'e}~MF Moura.
\newblock Adversarial geometry-aware human motion prediction.
\newblock In {\em Proceedings of the European Conference on Computer Vision
  (ECCV)}, pages 786--803, 2018.

\bibitem{guo2023back}
Wen Guo, Yuming Du, Xi Shen, Vincent Lepetit, Xavier Alameda-Pineda, and
  Francesc Moreno-Noguer.
\newblock Back to mlp: A simple baseline for human motion prediction.
\newblock In {\em Proceedings of the IEEE/CVF Winter Conference on Applications
  of Computer Vision}, pages 4809--4819, 2023.

\bibitem{henzinger2000theory}
Thomas~A Henzinger.
\newblock The theory of hybrid automata.
\newblock In {\em Verification of digital and hybrid systems}, pages 265--292.
  Springer, 2000.

\bibitem{hernandez2019human}
Alejandro Hernandez, Jurgen Gall, and Francesc Moreno-Noguer.
\newblock Human motion prediction via spatio-temporal inpainting.
\newblock In {\em Proceedings of the IEEE/CVF International Conference on
  Computer Vision}, pages 7134--7143, 2019.

\bibitem{hupkes2020compositionality}
Dieuwke Hupkes, Verna Dankers, Mathijs Mul, and Elia Bruni.
\newblock Compositionality decomposed: How do neural networks generalise?
\newblock {\em Journal of Artificial Intelligence Research}, 67:757--795, 2020.

\bibitem{jain2016structural}
Ashesh Jain, Amir~R Zamir, Silvio Savarese, and Ashutosh Saxena.
\newblock Structural-rnn: Deep learning on spatio-temporal graphs.
\newblock In {\em Proceedings of the ieee conference on computer vision and
  pattern recognition}, pages 5308--5317, 2016.

\bibitem{ji2021detecting}
Jingwei Ji, Rishi Desai, and Juan~Carlos Niebles.
\newblock Detecting human-object relationships in videos.
\newblock In {\em Proceedings of the IEEE/CVF International Conference on
  Computer Vision}, pages 8106--8116, 2021.

\bibitem{kundu2019bihmp}
Jogendra~Nath Kundu, Maharshi Gor, and R~Venkatesh Babu.
\newblock Bihmp-gan: Bidirectional 3d human motion prediction gan.
\newblock In {\em Proceedings of the AAAI conference on artificial
  intelligence}, volume~33, pages 8553--8560, 2019.

\bibitem{lecun2015deep}
Yann LeCun, Yoshua Bengio, and Geoffrey Hinton.
\newblock Deep learning.
\newblock {\em nature}, 521(7553):436--444, 2015.

\bibitem{li2020dynamic}
Maosen Li, Siheng Chen, Yangheng Zhao, Ya Zhang, Yanfeng Wang, and Qi Tian.
\newblock Dynamic multiscale graph neural networks for 3d skeleton based human
  motion prediction.
\newblock In {\em Proceedings of the IEEE/CVF Conference on Computer Vision and
  Pattern Recognition}, pages 214--223, 2020.

\bibitem{liu2020forecasting}
Miao Liu, Siyu Tang, Yin Li, and James~M Rehg.
\newblock Forecasting human-object interaction: joint prediction of motor
  attention and actions in first person video.
\newblock In {\em European Conference on Computer Vision}, pages 704--721.
  Springer, 2020.

\bibitem{liu2021motion}
Zhenguang Liu, Pengxiang Su, Shuang Wu, Xuanjing Shen, Haipeng Chen, Yanbin
  Hao, and Meng Wang.
\newblock Motion prediction using trajectory cues.
\newblock In {\em Proceedings of the IEEE/CVF international conference on
  computer vision}, pages 13299--13308, 2021.

\bibitem{ma2022progressively}
Tiezheng Ma, Yongwei Nie, Chengjiang Long, Qing Zhang, and Guiqing Li.
\newblock Progressively generating better initial guesses towards next stages
  for high-quality human motion prediction.
\newblock In {\em Proceedings of the IEEE/CVF Conference on Computer Vision and
  Pattern Recognition}, pages 6437--6446, 2022.

\bibitem{Mandery2015a}
Christian Mandery, \"Omer Terlemez, Martin Do, Nikolaus Vahrenkamp, and Tamim
  Asfour.
\newblock The kit whole-body human motion database.
\newblock In {\em International Conference on Advanced Robotics (ICAR)}, pages
  329--336, 2015.

\bibitem{mao2019learning}
Wei Mao, Miaomiao Liu, Mathieu Salzmann, and Hongdong Li.
\newblock Learning trajectory dependencies for human motion prediction.
\newblock In {\em Proceedings of the IEEE/CVF International Conference on
  Computer Vision}, pages 9489--9497, 2019.

\bibitem{martinez2017human}
Julieta Martinez, Michael~J Black, and Javier Romero.
\newblock On human motion prediction using recurrent neural networks.
\newblock In {\em Proceedings of the IEEE Conference on Computer Vision and
  Pattern Recognition}, pages 2891--2900, 2017.

\bibitem{morais2021learning}
Romero Morais, Vuong Le, Svetha Venkatesh, and Truyen Tran.
\newblock Learning asynchronous and sparse human-object interaction in videos.
\newblock In {\em Proceedings of the IEEE/CVF Conference on Computer Vision and
  Pattern Recognition}, pages 16041--16050, 2021.

\bibitem{nagarajan2019grounded}
Tushar Nagarajan, Christoph Feichtenhofer, and Kristen Grauman.
\newblock Grounded human-object interaction hotspots from video.
\newblock In {\em Proceedings of the IEEE/CVF International Conference on
  Computer Vision}, pages 8688--8697, 2019.

\bibitem{pavllo2018quaternet}
Dario Pavllo, David Grangier, and Michael Auli.
\newblock Quaternet: A quaternion-based recurrent model for human motion.
\newblock {\em arXiv preprint arXiv:1805.06485}, 2018.

\bibitem{qi2018learning}
Siyuan Qi, Wenguan Wang, Baoxiong Jia, Jianbing Shen, and Song-Chun Zhu.
\newblock Learning human-object interactions by graph parsing neural networks.
\newblock In {\em Proceedings of the European Conference on Computer Vision
  (ECCV)}, pages 401--417, 2018.

\bibitem{sofianos2021space}
Theodoros Sofianos, Alessio Sampieri, Luca Franco, and Fabio Galasso.
\newblock Space-time-separable graph convolutional network for pose
  forecasting.
\newblock In {\em Proceedings of the IEEE/CVF International Conference on
  Computer Vision}, pages 11209--11218, 2021.

\bibitem{van2001transients}
Lou Van~der Sluis.
\newblock {\em Transients in power systems}.
\newblock John Wiley \& Sons Ltd, 2001.

\bibitem{vaswani2017attention}
Ashish Vaswani, Noam Shazeer, Niki Parmar, Jakob Uszkoreit, Llion Jones,
  Aidan~N Gomez, {\L}ukasz Kaiser, and Illia Polosukhin.
\newblock Attention is all you need.
\newblock {\em Advances in neural information processing systems}, 30, 2017.

\bibitem{velivckovic2017graph}
Petar Veli{\v{c}}kovi{\'c}, Guillem Cucurull, Arantxa Casanova, Adriana Romero,
  Pietro Lio, and Yoshua Bengio.
\newblock Graph attention networks.
\newblock {\em arXiv preprint arXiv:1710.10903}, 2017.

\bibitem{wang2019heterogeneous}
Xiao Wang, Houye Ji, Chuan Shi, Bai Wang, Yanfang Ye, Peng Cui, and Philip~S
  Yu.
\newblock Heterogeneous graph attention network.
\newblock In {\em The World Wide Web Conference}, pages 2022--2032, 2019.

\bibitem{zhang2021we}
Yan Zhang, Michael~J Black, and Siyu Tang.
\newblock We are more than our joints: Predicting how 3d bodies move.
\newblock In {\em Proceedings of the IEEE/CVF Conference on Computer Vision and
  Pattern Recognition}, pages 3372--3382, 2021.

\end{thebibliography}
